\documentclass{article}

\PassOptionsToPackage{numbers}{natbib}
\usepackage[preprint]{neurips_2026}
\usepackage[utf8]{inputenc} 
\usepackage[T1]{fontenc}    
\usepackage{hyperref}       
\usepackage{url}            
\usepackage{booktabs}       
\usepackage{amsfonts}       
\usepackage{nicefrac}       
\usepackage{microtype}      
\usepackage{xcolor}         
\usepackage{amsmath}
\usepackage{graphicx}
\usepackage{bm}
\usepackage{enumitem}
\usepackage{subcaption}

\usepackage{booktabs, multirow} 
\usepackage{soul}
\usepackage{xcolor,colortbl} 
\usepackage{changepage,threeparttable} 

\newcommand{\Dl}{\mathcal{D}_l}
\newcommand{\Da}{\mathcal{D}_a}
\newcommand{\De}{\mathcal{D}_e}
\newcommand{\Ma}{\mathcal{M}_A}
\newcommand{\Mb}{\mathcal{M}_B}

\setlist{leftmargin=2em}

\title{Changing Modalities: Adapting Remote Sensing Models to New Satellites and Sensors}

\author{
Tim G. Zhou\textsuperscript{1,2} \quad
Anthony Fuller\textsuperscript{3,2} \quad
Geoff Pleiss\textsuperscript{1,2} \quad
Evan Shelhamer\textsuperscript{1,2} \\
\textsuperscript{1}University of British Columbia \quad
\textsuperscript{2}Vector Institute \quad
\textsuperscript{3}Carleton University \\
}

\begin{document}

\maketitle

\begin{abstract}
Machine learning models for remote sensing are trained and deployed on a static set of modalities.
However, as we equip newer satellites with novel sensors and retire old ones, practitioners may wish to deploy an existing model on a substitution, superset, or subset of modalities with minimal retraining given data availability or practical computational constraints.
We study the setting of updating existing models to changing modalities and identify three main scenarios: Modality Transfer (substitution), Addition (superset), and Peeking (subset). 
We propose DeluluNet, an architecture with modular components for all three changing modality scenarios.
DeluluNet is trained end-to-end, learning a multi-modal model from a unimodal 
teacher and \emph{un}labeled multimodal data via modality hallucination—predicting 
missing modality representations from those that are present.
As a result, DeluluNet can keep predicting even when input modalities change, 
providing a practical alternative to re-labeling and re-training in a changing world.
\end{abstract}

\section{Introduction: Modalities can Change, so Models Must Change}
Earth observation (EO) satellites continually collect vast amounts of \emph{diverse} data across modalities (optical \citep{Goward2017LandsatsEL}, radar \citep{ESA_sentinel-1}, lidar \citep{MARKUS2017260}).
The quantity and diversity of data make machine learning for remote sensing promising but challenging.
Satellite missions are regularly updated—sensors added or retired, new missions launched at higher resolutions, and formerly free data paywalled.
It is difficult to make use of these new modalities with existing remote sensing models without more training and annotations.
This poses a major challenge since models fail on modalities \emph{not} seen during training, e.g., a model trained on Sentinel-2's 12-band imagery cannot make predictions on the newest hyperspectral sensor's hundreds of bands \citep{Canada_2019}.
Ideally, a model trained on labeled data from existing satellites could be \emph{adapted} to make use of the new modality, without labeling new data or training from scratch.
We identify three scenarios under which models must adapt to new modalities:
\begin{enumerate}
    \item \emph{Modality Transfer}: switching a model from an old modality to a new one while maintaining predictions. For example, when Sentinel-2 \citep{DRUSCH201225} became available, many sought to convert LandSAT \citep{Goward2017LandsatsEL} models to take advantage of its higher spatial, spectral, and temporal resolution.
    \item \emph{Modality Addition}: incorporating a new modality into a model's predictions. When Sentinel-1 SAR data became available \citep{ESA_sentinel-1}, it was combined with existing optical imagery from LandSAT—a process that required re-labeling, but could be avoided with label-free modality addition.
    \item \emph{Modality Peeking}: improving a model's predictions on an old modality through limited observation of a new one. Global hyperspectral imagery \citep{Canada_2019} is rarely freely available, but is rich enough that a model could peek at it to learn fine-grained spectral patterns and leverage the knowledge acquired to improve predictions on standard optical data.
\end{enumerate}

These changing modality settings have been relatively unexplored in the literature.
Existing machine learning techniques have limited applicability: a given method may work for one of the three settings but not the others.
A key contribution of our work is recognizing that transfer, addition, and peeking share similarities that are better served by a unified approach.
We thus propose \textbf{DeluluNet}, an architecture with modular uni-modal and multi-modal components that adapts an existing predictive model using unlabeled multi-modal data.
DeluluNet leverages its multi-modal components even when test data is incomplete, imputing missing modality representations by predicting from the modalities that are present. 
We evaluate DeluluNet on two types of RS modality change: cross-band scenarios within Sentinel-2, and cross-satellite scenarios between Sentinel-1 and Sentinel-2. These experiments demonstrate its efficacy across transfer, addition, and peeking, covering changes in both channel bands and satellites, and all without requiring any labels on the new modality data.

\begin{figure*}[t]
    \centering
    \begin{minipage}[c]{0.55\textwidth}
        \includegraphics[width=1\textwidth]{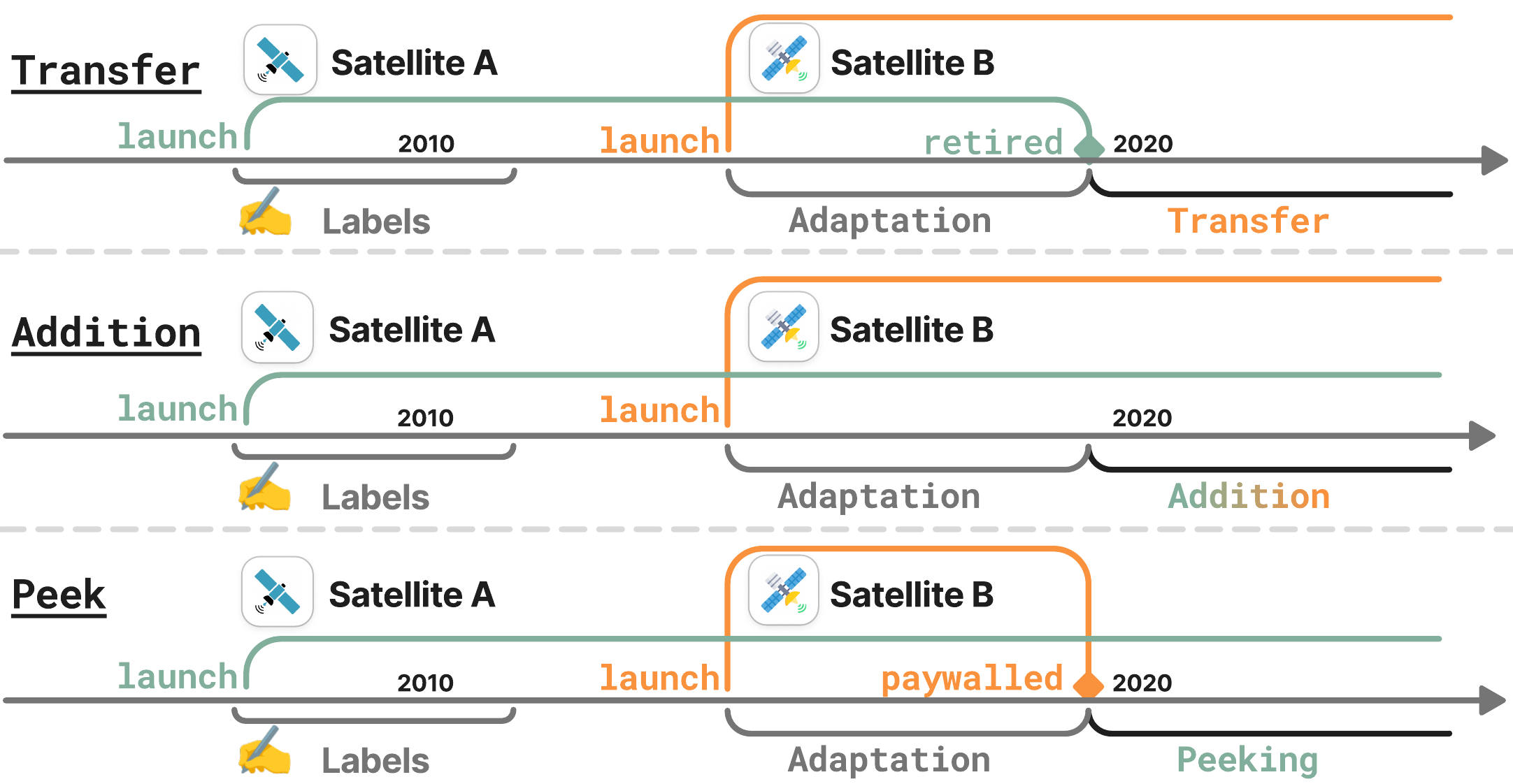}
    \end{minipage}
    \hfill
    \begin{minipage}[c]{0.39\textwidth}
        \caption{\textbf{Changing modalities:} When satellites are launched, paywalled, or retired, models must adapt to new sensor data without re-labeling. We consider three scenarios: \emph{modality transfer} (predict using only new Satellite B data), \emph{modality peeking} (predict using only old Satellite A data, informed by B), and \emph{modality addition} (predict using both A and B).}
        \label{fig:setting}
    \end{minipage}
\end{figure*}

\section{Setting: Changing Modalities}
\label{sec:changing-modalities}

\textbf{Data splits and modalities.}
Consider a predictive task with labels $\mathcal Y$ and two possible input modalities: a \emph{partially-labeled modality $\Ma$} and an \emph{unlabeled modality $\Mb$}. Data for learning is then divided into three splits:
\begin{enumerate}
    \item \textbf{Labeled split $\Dl \subset (\Ma \times \mathcal{Y})$} for the initial supervised training.
    \item \textbf{Adaptation split $\Da \subset (\Ma \times \Mb)$} for unsupervised adaptation.
    \item \textbf{Evaluation split $\De$} contains either or both of $\Ma$ and $\Mb$, depending on the scenario.
\end{enumerate}

\textbf{Scenarios.}
Three main scenarios stem naturally from the changing modalities setting, each corresponding to a different evaluation split:
\begin{enumerate}
\item \textbf{Modality Transfer $\De \subset \Mb$}: The adapted model makes predictions using $\Mb$ alone, having transferred predictive ability from $\Ma$ to $\Mb$ without supervision.
\item \textbf{Modality Addition $\De \subset (\Ma \cup \Mb)$}: The adapted model makes predictions using $\Ma$ and $\Mb$ jointly, leveraging the two modalities jointly.
\item \textbf{Modality Peeking $\De \subset \Ma$}: The adapted model continues to make predictions using $\Ma$ alone, but with additional knowledge from observing unlabeled $\Mb$ during adaptation.
\end{enumerate}

In all scenarios, we start with a uni-modal predictive model $f_0$ trained with supervision on the labeled split $\Dl$, adapt $f_0$ on $\Da$ into a multi-modal DeluluNet, and evaluate on $\De$ (see Figure~\ref{fig:setting}).

\section{Background: Foundation Models, Scenarios, and Remote Sensing}
Prior work adapts models on unlabeled data, but assumes a constant input modality.
We review why existing models and methods are insufficient for the transfer, addition, and peeking scenarios.

\paragraph{Existing remote sensing foundation models cannot predict well on unseen modalities.}
To handle RS data from a variety of satellite modalities, RS Foundation Models (RSFMs) are pretrained on large multi-modal datasets and extract useful information across diverse downstream use cases \citep{lacoste2023geo}. 
Most recently, the field has seen a wave of remote sensing foundation models (RSFMs) \citep{fuller2023croma,10641637,pmlr-v267-tseng25a,herzog2025olmoearthstablelatentimage,labatie2025maestromaskedautoencodersmultimodal} following the success of language \citep{radford2021learning} and vision foundation models \citep{simeoni2025dinov3}. 
These RSFMs are powerful and \emph{multi-modal}, taking in various modalities of remote sensing, such as multi-spectral optical data (e.g. Sentinel-2, LandSAT 8\&9) and Synthetic Aperture Radar data (e.g. Sentinel-1).
Many such models utilize some modality-aware masked image modeling objectives \citep{mizrahi20234m} to enable usability when some modalities are absent \citep{cong2022satmae,herzog2025olmoearthstablelatentimage,zhang2025skysense}, but they are constrained to a set of modalities seen during training. 
As a result, they require costly re-pretraining, relabeling, and re-finetuning when sensor modalities change \citep{tamazyan2025geocrossbench}. 
While ``sensor-agnostic'' remote sensing models \citep{xiong2024neural,waldmann2025panopticon} are trained to handle arbitrary numbers of spectral channels at any wavelength, and thus are suitable for any satellite modality in theory, they become specialized to the labeled modality they are finetuned on.
They cannot readily make use of an adaptation set to update a finetuned model given a new modality.

\paragraph{The three scenarios are scattered in different sub-fields of deep learning.}
Variants of the three changing modalities scenarios have been investigated in different sub-fields of deep learning.
Below we discuss methods that, although not designed for changing modalities in general, can nevertheless be applied to one of the scenarios.

\begin{itemize}
\item \emph{Knowledge Distillation}  \citep{Bucila2006Model,hinton2015distilling} studies transferring knowledge from one deep network to another, typically assuming a shared input modality between teacher and student. The transfer scenario we propose is an under-explored special case of knowledge transfer, where we transfer predictive power on the old modality to the new, unlabeled modality. The closest sub-branch of KD to our \textit{modality transfer} setting is the cross-modality distillation \citep{gupta2016cross,Zhao_2018_CVPR,zhou2023unidistill}, which assumes labeled multimodal data, enabling a supervised hard loss term in addition to soft distillation loss.

\item \emph{Semi-Supervised Learning} (Semi-SL) \citep{blum1998combining} combines labeled data and unlabeled data, often leveraging one or several of entropy minimization \citep{Grandvalet2004advances}, pseudo-labeling \citep{lee2013pseudo,cascante2021curriculum} and consistency regularization \citep{sohn2020fixmatch,wang2023freematch,berthelot2019mixmatch}. Semi-SL, similar to \emph{modality peeking}, learns from labeled and unlabeled data splits, but assumes that labeled and unlabeled data come from the same modality. 

\item\emph{Multimodal Knowledge Expansion} (MKE) \citep{xue2021multimodal} sits on the intersection of knowledge distillation and Semi-SL, and corresponds to our modality addition setting, learning a multimodal student from a unimodal teacher. 
Our setting is more general by additionally considering transfer and peeking scenarios that flexibly predict with any subsets of the considered modalities.
\end{itemize}

No existing method addresses all three scenarios jointly.
Each sub-field has developed solutions in isolation, and none are designed for the label-free setting of changing modalities that arises naturally in remote sensing. 
DeluluNet bridges this gap with a single unified framework.

\paragraph{Adjacent fields assume fixed modalities.}
\textit{Continual Learning} is a related setting which focuses on continuously acquiring new knowledge while preserving learned knowledge through Task/Class/Domain incremental learning \citep{MCCLOSKEY1989109,van2022three,lu2025rethinking}; however, it assumes both labeled data and a fixed modality.
\emph{Domain Adaptation} \citep{ganin2016domain,sun2016deep} focuses on transferring knowledge from a labeled source domain to a different and unlabeled target domain; 
however it also assumes the input modality remains unchanged across source/train and target/test domains.

\paragraph{Modality change is broadly relevant in applied machine learning—especially in remote sensing.}
This work focuses on changing modalities for remote sensing, where it is relevant due to the abundance of unlabeled multi-modal data from orbiting satellites that continually capture data.
However, the changing modalities setting is relevant in many applied machine learning fields.
Modality transfer could be useful for adapting existing models to new medical devices \citep{phillips2020teacher}, modality addition could be useful for adding new sensors for autonomous driving \citep{hegde2025modality}, and modality peeking could leverage high-cost imaging such as CT scans during training and still deploy with only more accessible alternatives such as chest X-rays in resource-constrained settings \citep{cao2024bootstrapping}.

\begin{figure*}[t]
    \centering
    \includegraphics[width=1\textwidth]{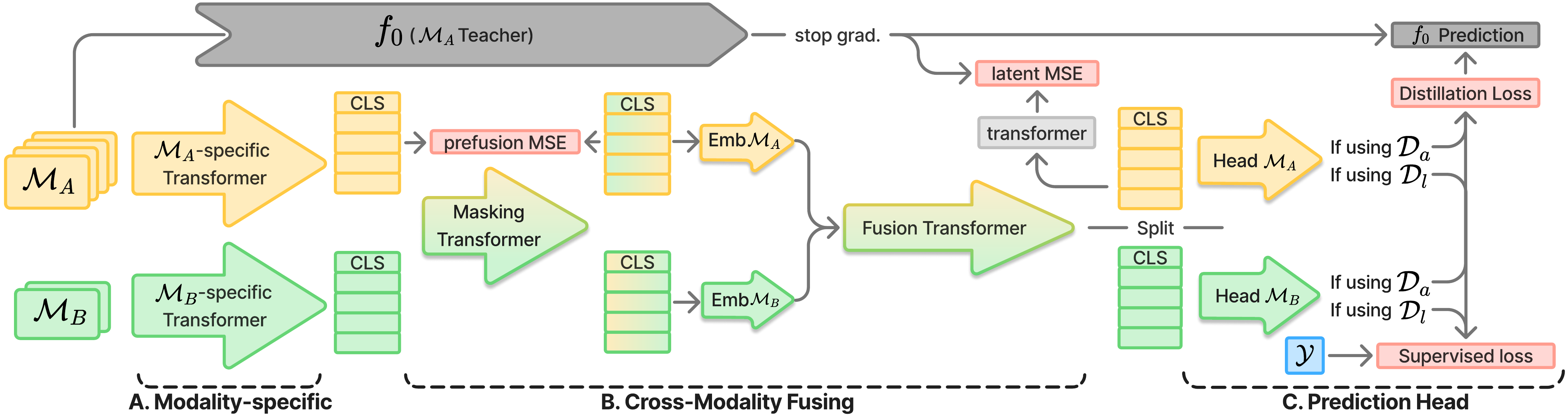}
    \caption{\textbf{DeluluNet Architecture and Training Strategy.} 
    \emph{A.} Shallow modality-specific transformers extract low-level information from available modalities; 
    \emph{B.} Masking and fusion transformers facilitate cross-modality fusion through masked image modeling, and enables prediction with incomplete input modalities at test time;
    \emph{C.} Predictor heads learning switches between distilling from uni-modal $f_0$ teacher on $\Da$ and learning from real labels from $\Dl$.}
    \label{fig:delulu_architecture}
\end{figure*}
\begin{figure*}[t]
    \centering
    \begin{minipage}[c]{0.49\textwidth}
        \includegraphics[width=\linewidth]{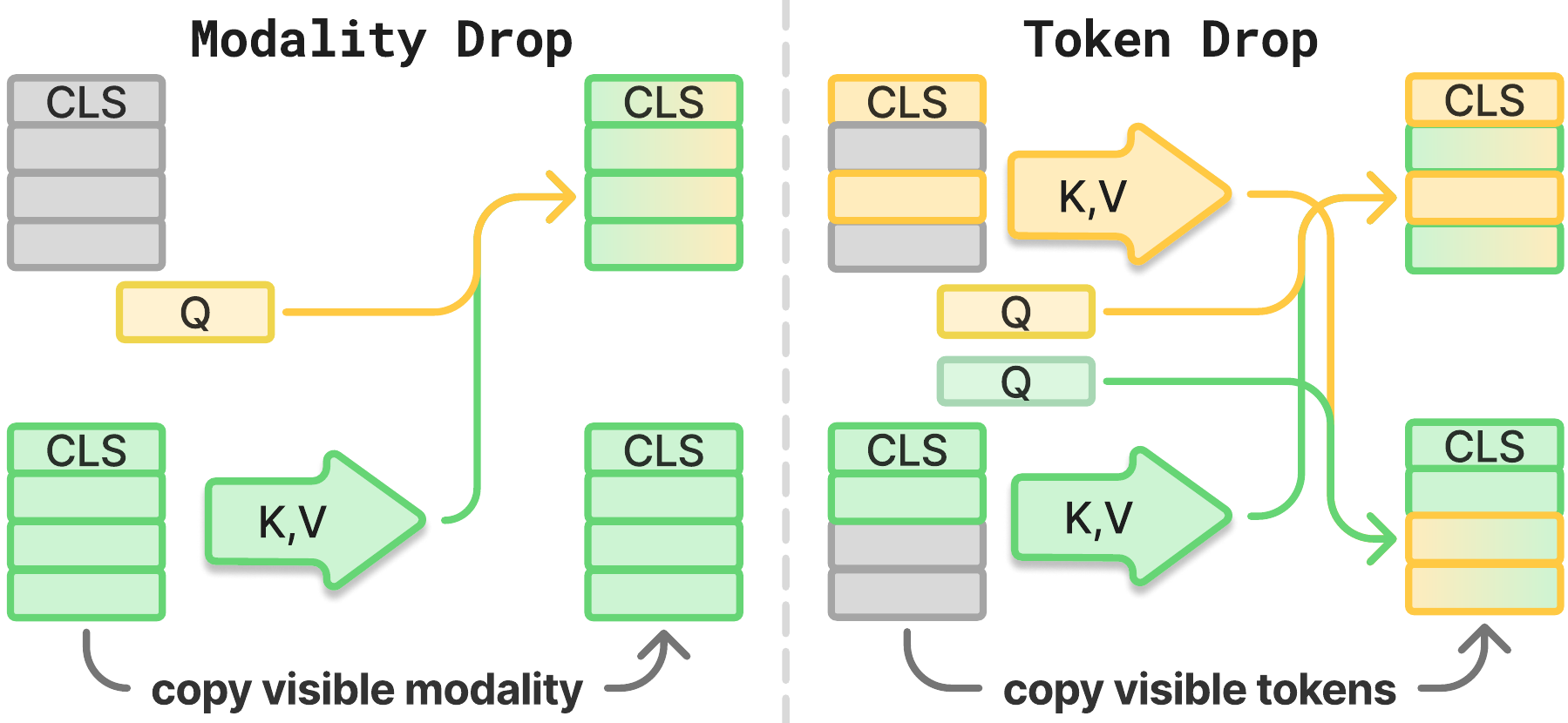}
    \end{minipage}
    \hfill
    \begin{minipage}[c]{0.48\textwidth}
        \caption{\textbf{Masking transformer cross-attention under modality-drop (left) and token-drop (right).}
        In both cases, each modality's learned query $Q$ attends to visible tokens from the other modality as $K,V$ to predict masked tokens.
        Transfer and peeking at test time naturally follow the modality-drop path, while addition requires no masking transformer.}
        \label{fig:masking_transformer}
    \end{minipage}
\end{figure*}

\section{Method: Modality Hallucination (Delulu is the Solulu)}
\label{Sec:method}
DeluluNet operates under the changing-modality setting by augmenting an existing 
uni-modal model with modular components for inter- and intra-modality learning, 
handling transfer, addition, and peeking within a single end-to-end framework. 
Starting from a ViT \citep{dosovitskiy2021an} on $\Ma$-input as the predictive model $f_0$ trained on labeled 
$\Dl$, DeluluNet introduces three modular components to support feature extraction 
and prediction with the new modality $\Mb$ (Figure~\ref{fig:delulu_architecture}):
\begin{enumerate}
    \item \emph{Modality-Specific Extraction:} a randomly-initialized $\Mb$-specific 
    tokenizer and $f_0$-initialized transformer layers for modality-specific representation of $\Mb$.
    \item \emph{Cross-Modality Fusion:} randomly-initialized modality embeddings 
    $\text{Emb}_{\Mb}$ and masking transformers that predict cross-modal features between $\Ma$ and $\Mb$.
    \item \emph{Predictor Head:} a task-specific $\text{Head}_{\Mb}$ that predicts from $\Mb$ representations.
\end{enumerate}
At test time, when both modalities are available (addition), each passes through its 
respective extractor before cross-modality fusion. When only one modality is present 
(transfer or peeking), the masking transformers hallucinate intermediate 
representations for the missing modality, enabling the model to treat it as 
multi-modal input. We next describe how these components are trained from unlabeled $\Da$.

\subsection{Learning Features and Predictions From Unlabeled Modalities.}

DeluluNet is trained with the same end-to-end strategy for all three scenarios, with losses at different depths supervising each modular component.

\paragraph{Modality-specific feature learning.}
To extract low-level features tailored to each modality's characteristics, DeluluNet employs shallow modality-specific transformers: patch embedders that project each modality's inputs into a shared dimension $d$, followed by $n_{\mathrm{ms}}$ transformer blocks.
This design is common among multimodal RSFMs, which vary in fusion depth from early-fusion ($n_{\mathrm{ms}} = 0$) \citep{tseng2025galileo,herzog2025olmoearthstablelatentimage} to mid-fusion \citep{astruc2024omnisat} and dynamic fusion \citep{labatie2025maestromaskedautoencodersmultimodal}; we fix $n_{\mathrm{ms}}=3$ and ablate fusion depth in Section~\ref{sec:masking_transformer_analysis}.
The $\Ma$-specific components are initialized from $f_0$, while $\Mb$-specific components are randomly initialized.

\paragraph{Cross-modality hallucination via masking transformers.}
To facilitate cross-modal feature learning and prepare DeluluNet for missing modalities at test time, we introduce \textbf{masking transformers} inspired by masked image modeling \citep{he2022masked,cong2022satmae,tseng2025galileo,herzog2025olmoearthstablelatentimage} that hallucinate masked representations from visible ones.
For each mini-batch, we adopt masking at two granularities (Figure~\ref{fig:masking_transformer}):
\emph{token-level masking} randomly masks tokens across all modalities, providing cross-modality learning signals when both modalities are partially available;
\emph{modality-level masking} treats an entire modality as absent, preparing the model for the missing-modality inference.
The masking transformer imputes masked tokens and modalities from visible ones via cross-attention: a learned per-modality query attends to visible tokens from other modalities, with one CLS and one patch token, each of dimension $d$. 
In forward passes, the patch token is broadcast to $(n, d)$ and augmented with $f_0$'s position encodings, keeping the design parameter-efficient.
A \textsc{prefusion mse} loss directly supervises this hallucination by encouraging predicted CLS and patch tokens to match the true tokens of the masked modality.

\paragraph{Modality fusion learning.}
To combine features across modalities and teach DeluluNet to leverage hallucinated representations, a fusion transformer operates on the concatenation of all modality tokens via standard transformer blocks, allowing tokens from different modalities to attend to each other.
Since transformers are applicable to any sequence of tokens sharing a hidden dimension, this naturally accommodates both real and hallucinated representations, so the fusion transformer always receives a full multi-modal input regardless of which modalities are present at test time.
To anchor fused representations in a task-relevant feature space, a \textsc{latent mse} loss is applied between the fusion transformer output and $f_0$'s frozen penultimate representations via a 1-layer transformer decoder.

\paragraph{Prediction head learning.}
To ensure accurate predictions that incorporate features from both modalities, we alternate mini-batches between the multi-modal unlabeled $\Da$ and the uni-modal labeled dataset $\Dl$.
As a result, DeluluNet receives two sources of supervision for predictor heads, a technique we refer to as \textbf{batch mixing}, made possible by the masking transformer:
\begin{itemize}
    \item On batches from the \emph{unlabeled $\Da$}, DeluluNet applies token- and modality-level masking and distills \citep{hinton2015distilling} from the predictions made by the uni-modal teacher $f_0$ using a \textsc{distillation loss}. 
    The unlabeled batches allow learning from pseudo-labels using real multi-modal inputs.
    \item On batches from the \emph{labeled $\Dl$}, DeluluNet treats these as inputs where $\Mb$ is masked, and a supervised \textsc{cross-entropy loss} is applied to predictions.
    The labeled batches allow learning from real labels using pseudo multi-modal inputs from the masking transformer.
\end{itemize}

We ablate masking transformers, MSE losses, and batch mixing in Section~\ref{sec:masking_transformer_analysis}, showing each is crucial for DeluluNet's performance across scenarios.
Further architecture details are in Appendix~\ref{apx:architecture}.

\section{Experiments across Changing Bands and Satellites}
\label{sec:experiment}

\paragraph{Datasets and bands.} 
We evaluate DeluluNet across single-label classification, multi-label classification, and semantic segmentation with remote sensing benchmarks.
We study both cross-band modality changes using channel groups within Sentinel 2 (e.g., RGB to Vegetation Red-Edge bands), and cross-satellite modality changes between Sentinel-1 and Sentinel-2 (e.g., Synthetic Aperture Radar data to Multi-Spectral Images).
\textbf{EuroSAT} \citep{helber2019eurosat} is the canonical 10-way \emph{classification} dataset with 13 spectral bands from the Sentinel-2 satellites. 
We select two of the major groups of bands from EuroSAT, namely RGB (B-02,B-03,B-04) and VRE (B-05, B-06, B-07) to evaluate DeluluNet's performance for cross-band modality change, and present more results on cross-band transfer, peeking, and addition can be found in Appendix~\ref{apx:eurosat}.
\textbf{reBEN} \citep{clasen2025reben} is the latest update of the classic BigEarthNet dataset \citep{sumbul2019bigearthnet}, a large European multi-modal (2-band Sentinel-1 and 12-band Sentinel-2) dataset for \emph{multi-label Land Cover Classification}. 
We use reBEN from Geo-Bench-2 \citep{simumba2025geo} to study DeluluNet's ability to adapt to an entirely new satellite without labels.
\textbf{DFC2020} \citep{9369830} is a 8-class land-cover \emph{semantic segmentation} benchmark, multi-modal with Sentinel-1 (2 bands) and Sentinel-2 (13 bands) data. We use it to study DeluluNet's ability to adapt dense prediction tasks.
For all benchmarks, we randomly split the official training data in half: \textit{labeled uni-modal} $\Dl$ and \textit{unlabeled multi-modal} $\Da$.

\paragraph{Architecture.}
We train supervised \textbf{ViT-Base} \citep{dosovitskiy2021an} on the labeled $\Dl$, initialized with pre-trained \emph{DINO v3} \citep{simeoni2025dinov3} weights for the modality-fusion blocks and uni-modal transformers. 
These supervised models serve as the initial uni-modal $f_0$ for DeluluNet.
We contextualize the efficacy of DeluluNet by fully fine-tuning the recent and high-accuracy multi-modal RSFMs OlmoEarth (Base) \citep{herzog2025olmoearthstablelatentimage} and Panopticon (Base) \citep{waldmann2025panopticon} on labeled uni- and multi-modal data as oracles. 
These RSFMs represent the most competitive performance achievable when models are pre-trained with massive multi-modal data and fine-tuned with labeled multi-modal inputs.

\paragraph{Optimization and hyper-parameter tuning.}
The main hyper-parameters to tune for DeluluNet include learning rate, weight decay, masking ratio (token and modality), and batch-mixing ratio. 
Since DeluluNet can handle all three scenarios with one end-to-end strategy, we separately tune hyper-parameters for the three scenarios using reBEN dataset with Sentinel-1 as starting modality and Sentinel 2 as new modality.
Validation scoring is also label-free for multi-modal evaluation via agreement-based metric detailed in Appendix~\ref{apx:validation}.
All DeluluNet experiments are trained with these hyper-parameters to avoid per-dataset-architecture-$M_A$-$M_B$ tuning.
Even though better performance is achievable through independent hyper-parameter tuning for each dataset and $\Ma$-$\Mb$ pairing, this simple and cheap approach is already effective across datasets, modalities, and architectures.

\begin{table}[t]
\centering
\caption{\textbf{Modality Transfer} results show that DeluluNet transfers more effectively than knowledge distillation (KD) and transformed teacher matching (TTM), and at times even surpasses supervised fine-tuned DINOv3 (DiNO SFT) on $M_B$.}
\resizebox{\textwidth}{!}{%
  \begin{tabular}{cc | c c c c | c c c c | c}
\toprule
  \multicolumn{2}{c|}{\shortstack[c]{Dataset\\(metric)}} & \multicolumn{4}{c|}{\shortstack[c]{reBEN\\(mAP)}} & \multicolumn{4}{c|}{\shortstack[c]{DFC2020\\(mIoU)}} & \multicolumn{1}{c}{\shortstack[c]{EuroSAT\\(Acc)}} \\
  \midrule
  \multicolumn{2}{c|}{Start ($M_A$)} & \multicolumn{2}{c}{$\mathrm{S2}_{rgb}$} & \multicolumn{1}{c}{$\mathrm{S1}$} & \multicolumn{1}{c|}{$\mathrm{S2}$} & \multicolumn{2}{c}{$\mathrm{S2}_{rgb}$} & \multicolumn{1}{c}{$\mathrm{S1}$} & \multicolumn{1}{c|}{$\mathrm{S2}$} & \multicolumn{1}{c}{RGB} \\
  \multicolumn{2}{c|}{Transfer ($M_B$)} & $\mathrm{S1}$ & $\mathrm{S2}_{\neg rgb}$ & $\mathrm{S2}$ & $\mathrm{S1}$ & $\mathrm{S1}$ & $\mathrm{S2}_{\neg rgb}$ & $\mathrm{S2}$ & $\mathrm{S1}$ & VRE \\
  \midrule
  $f_0(M_A)$ & $f_0$ & 63.4 & 63.4 & 49.2 & 61.3 & 41.9 & 41.9 & 43.3 & 45.1 & 98.6 \\
  \midrule
  \multirow{2}{*}{Baselines} & KD & 48.0$\pm$2.3 & 57.4$\pm$0.9 & 48.8$\pm$0.7 & 49.3$\pm$0.0 & 43.2$\pm$0.6 & 42.4$\pm$2.2 & 35.6$\pm$2.3 & \textbf{45.6$\pm$0.8} & 88.6$\pm$1.3 \\
   & TTM & 47.4$\pm$2.3 & 56.7$\pm$0.5 & 49.3$\pm$0.4 & 48.7$\pm$2.1 & 43.3$\pm$1.5 & 44.9$\pm$2.2 & 40.0$\pm$1.3 & 45.2$\pm$0.5 & 89.5$\pm$1.0 \\
  Ours & Delulu & \textbf{50.8$\pm$0.3} & \textbf{62.5$\pm$1.2} & \textbf{52.1$\pm$0.2} & \textbf{49.9$\pm$0.3} & \textbf{45.6$\pm$0.9} & \textbf{50.6$\pm$1.4} & \textbf{48.2$\pm$0.6} & 41.8$\pm$0.0 & \textbf{96.4$\pm$0.2} \\
  \midrule
  \multirow{3}{*}{\shortstack[c]{Oracle\\($M_B$)}} & DINOv3 & \textcolor{gray}{49.2} & \textcolor{gray}{57.6} & \textcolor{gray}{61.3} & \textcolor{gray}{49.2} & \textcolor{gray}{43.3} & \textcolor{gray}{39.9} & \textcolor{gray}{45.1} & \textcolor{gray}{43.3} & \textcolor{gray}{90.7} \\
   & Panopticon & \textcolor{gray}{55.1} & \textcolor{gray}{61.4} & \textcolor{gray}{62.1} & \textcolor{gray}{55.1} & \textcolor{gray}{48.1} & \textcolor{gray}{48.5} & \textcolor{gray}{50.5} & \textcolor{gray}{48.1} & \textcolor{gray}{97.8} \\
   & OlmoEarth & \textcolor{gray}{54.1} & \textcolor{gray}{69.6} & \textcolor{gray}{65.1} & \textcolor{gray}{54.1} & \textcolor{gray}{47.4} & \textcolor{gray}{54.0} & \textcolor{gray}{52.1} & \textcolor{gray}{47.4} & \textcolor{gray}{96.4} \\
\bottomrule
\end{tabular}%
}
\label{tab:transfer}

\vspace{8pt}
\caption{\textbf{Modality Addition} results show that DeluluNet is able to extract and fuse inputs from both $\Ma$ and $\Mb$ through modality addition, not only surpassing multimodal knowledge expansion (MKE) and DINO-init $M_A$ $f_0$, but sometimes even supervised RSFMs.}
\resizebox{\textwidth}{!}{%
  \begin{tabular}{cc | c c c c | c c c c | c}
\toprule
  \multicolumn{2}{c|}{\shortstack[c]{Dataset\\(metric)}} & \multicolumn{4}{c|}{\shortstack[c]{reBEN\\(mAP)}} & \multicolumn{4}{c|}{\shortstack[c]{DFC2020\\(mIoU)}} & \multicolumn{1}{c}{\shortstack[c]{EuroSAT\\(Acc)}} \\
  \midrule
  \multicolumn{2}{c|}{Start ($M_A$)} & \multicolumn{2}{c}{$\mathrm{S2}_{rgb}$} & \multicolumn{1}{c}{$\mathrm{S1}$} & \multicolumn{1}{c|}{$\mathrm{S2}$} & \multicolumn{2}{c}{$\mathrm{S2}_{rgb}$} & \multicolumn{1}{c}{$\mathrm{S1}$} & \multicolumn{1}{c|}{$\mathrm{S2}$} & \multicolumn{1}{c}{RGB} \\
  \multicolumn{2}{c|}{New ($M_B$)} & $\mathrm{S1}$ & $\mathrm{S2}_{\neg rgb}$ & $\mathrm{S2}$ & $\mathrm{S1}$ & $\mathrm{S1}$ & $\mathrm{S2}_{\neg rgb}$ & $\mathrm{S2}$ & $\mathrm{S1}$ & VRE \\
  \midrule
  $f_0(M_A)$ & $f_0$ & 63.4 & 63.4 & 49.2 & 61.3 & 41.9 & 41.9 & 43.3 & 45.1 & 98.6 \\
  Baselines & MKE & 62.2$\pm$0.6 & 61.9$\pm$0.1 & 49.3$\pm$0.5 & 60.8$\pm$0.2 & 42.2$\pm$4.4 & 39.1$\pm$2.7 & \textbf{43.4$\pm$3.3} & 38.0$\pm$0.7 & 97.5$\pm$0.9 \\
  Ours & Delulu & \textbf{64.7$\pm$0.5} & \textbf{63.5$\pm$0.2} & \textbf{49.8$\pm$0.4} & \textbf{61.4$\pm$0.6} & \textbf{47.3$\pm$0.9} & \textbf{47.0$\pm$2.1} & 42.8$\pm$0.9 & \textbf{46.9$\pm$0.8} & \textbf{98.7$\pm$0.1} \\
  \midrule
  \multirow{3}{*}{\shortstack[c]{Oracle\\($M_A$+$M_B$)}} & DINOv3 & \textcolor{gray}{63.5} & \textcolor{gray}{63.7} & \textcolor{gray}{61.5} & \textcolor{gray}{61.5} & \textcolor{gray}{45.0} & \textcolor{gray}{49.5} & \textcolor{gray}{39.4} & \textcolor{gray}{39.4} & \textcolor{gray}{98.6} \\
   & Panopticon & \textcolor{gray}{63.4} & \textcolor{gray}{62.5} & \textcolor{gray}{63.4} & \textcolor{gray}{63.4} & \textcolor{gray}{52.3} & \textcolor{gray}{50.8} & \textcolor{gray}{52.3} & \textcolor{gray}{52.3} & \textcolor{gray}{98.7} \\
   & OlmoEarth & \textcolor{gray}{64.6} & \textcolor{gray}{65.1} & \textcolor{gray}{64.6} & \textcolor{gray}{64.6} & \textcolor{gray}{55.5} & \textcolor{gray}{52.1} & \textcolor{gray}{55.5} & \textcolor{gray}{55.5} & \textcolor{gray}{98.3} \\
\bottomrule
\end{tabular}%
}
\label{tab:addition}

\vspace{8pt}
\caption{\textbf{Modality Peeking} results show that DeluluNet is able to learn from observations of unlabeled and paired $\Ma$ and $\Mb$ data to make better predictions on $\Ma$, consistently surpassing Semi-SL baseline and DINO-init $M_A$ $f_0$.}
\resizebox{\textwidth}{!}{%
  \begin{tabular}{cc | c c c c | c c c c | c}
\toprule
  \multicolumn{2}{c|}{\shortstack[c]{Dataset\\(metric)}} & \multicolumn{4}{c|}{\shortstack[c]{reBEN\\(mAP)}} & \multicolumn{4}{c|}{\shortstack[c]{DFC2020\\(mIoU)}} & \multicolumn{1}{c}{\shortstack[c]{EuroSAT\\(Acc)}} \\
  \midrule
  \multicolumn{2}{c|}{Start ($M_A$)} & \multicolumn{2}{c}{$\mathrm{S2}_{rgb}$} & \multicolumn{1}{c}{$\mathrm{S1}$} & \multicolumn{1}{c|}{$\mathrm{S2}$} & \multicolumn{2}{c}{$\mathrm{S2}_{rgb}$} & \multicolumn{1}{c}{$\mathrm{S1}$} & \multicolumn{1}{c|}{$\mathrm{S2}$} & \multicolumn{1}{c}{RGB} \\
  \multicolumn{2}{c|}{New ($M_B$)} & $\mathrm{S1}$ & $\mathrm{S2}_{\neg rgb}$ & $\mathrm{S2}$ & $\mathrm{S1}$ & $\mathrm{S1}$ & $\mathrm{S2}_{\neg rgb}$ & $\mathrm{S2}$ & $\mathrm{S1}$ & VRE \\
  \midrule
  $f_0(M_A)$ & $f_0$ & 63.4 & 63.4 & 49.2 & 61.3 & 41.9 & 41.9 & 43.3 & \textbf{45.1} & 98.6 \\
  Baselines & MixMatch & 52.1$\pm$6.4 & 52.1$\pm$6.4 & 47.7$\pm$2.2 & 59.5$\pm$2.3 & 39.1$\pm$3.4 & 39.1$\pm$3.4 & 35.4$\pm$2.8 & 40.1$\pm$3.7 & \textbf{99.0$\pm$0.1} \\
  Ours & Delulu & \textbf{64.5$\pm$1.2} & \textbf{63.6$\pm$0.7} & \textbf{49.4$\pm$0.3} & \textbf{62.5$\pm$0.6} & \textbf{43.9$\pm$2.7} & \textbf{46.7$\pm$1.3} & \textbf{43.5$\pm$0.9} & 42.1$\pm$0.7 & 98.5$\pm$0.1 \\
  \midrule
  \multirow{2}{*}{\shortstack[c]{Oracle\\($M_A$)}} & Panopticon & \textcolor{gray}{60.0} & \textcolor{gray}{60.0} & \textcolor{gray}{55.1} & \textcolor{gray}{62.1} & \textcolor{gray}{45.5} & \textcolor{gray}{45.5} & \textcolor{gray}{48.1} & \textcolor{gray}{50.5} & \textcolor{gray}{98.4} \\
   & OlmoEarth & \textcolor{gray}{59.7} & \textcolor{gray}{59.7} & \textcolor{gray}{54.1} & \textcolor{gray}{65.1} & \textcolor{gray}{49.7} & \textcolor{gray}{49.7} & \textcolor{gray}{47.4} & \textcolor{gray}{52.1} & \textcolor{gray}{97.7} \\
\bottomrule
\end{tabular}%
}
\label{tab:peek}
\end{table}


\paragraph{Modality transfer results.}
We compare DeluluNet in modality transfer scenario to canonical knowledge distillation (KD) \citep{hinton2015distilling} and the recent Transformed Teacher Matching (TTM) \citep{zheng2024knowledge} as label-free baselines, and additionally fine-tune fully supervised DINOv3, Panopticon, and OlmoEarth as label-oracles.
Table~\ref{tab:transfer} shows that DeluluNet not only consistently transfers more effectively than KD and TTM in the changing modality setting, but often surpasses the DINOv3-init supervised label-oracle models.
This is particularly true when the starting modality is RGB or the RGB bands from Sentinel-2—these are consistent with the DINOv3's RGB pretraining, providing a stronger starting point for transfer. 
For example, when transferring from $\text{S2}_{rgb}$ to $\text{S2}_{\neg rgb}$ (the non-rgb bands from Sentinel-2), Delulu is able to reach 62.5 mAP on reBEN, and 50.6 mIoU on DFC2020, a respective $+4.9$ and $+10.7$ improvement compared to a label-oracle fine-tuned DINO model, and in both cases reaching performance close to those of fine-tuned RSFMs.

\paragraph{Modality addition results.}
We compare DeluluNet in modality addition scenario to Multimodal Knowledge Expansion (MKE) \citep{xue2021multimodal}, the natural baseline for modality addition which learns a multi-modal model from a uni-modal teacher, aiming to make better predictions.
We found that while MKE can sometimes outperform the uni-modal $f_0$, DeluluNet provides more consistent and sizable improvements. 
Table~\ref{tab:addition} shows that across all modality settings we tested, DeluluNet always improves performance on $f_0$, suggesting that it is effective at improving predictive performance when new complementary modalities become available.
For example, when adding Sentinel-1 inputs to a $\text{S2}_{rgb}$ $f_0$, DeluluNet is able to improve mAP on reBEN by $1.3\%$, outperforming supervised DINOv3 oracle and even achieving performance comparable to supervised fine-tuned RSFM oracles.

\paragraph{Modality peeking results.}
We compare DeluluNet in modality peeking scenario to the popular Semi-SL method MixMatch \citep{berthelot2019mixmatch}, which trains a model on both labeled and unlabeled data using consistency regularization and pseudo labeling with various augmentations including Mixup \citep{zhang2018mixup}.
Table~\ref{tab:peek} shows that DeluluNet effectively learns from paired $\Ma,\Mb$ from $\Da$ and consistently improves on its uni-modal $f_0$ initialization.
For example, DeluluNet initialized with an $\text{S2}_{rgb}$ $f_0$ after peeking at $\text{S2}_{\neg rgb}$ achieves a 4.8\% improvement in mIoU on DFC2020, achieving performance close to RSFM orfacles supervised with actual labeled multi-modal data.

\begin{figure}[t]
    \centering
    \begin{minipage}[t]{0.6\textwidth}
        \vspace{0pt}
        \centering
        \resizebox{\textwidth}{!}{\begin{tabular}{lccc}
\toprule
Configuration & Transfer & Peek & Addition \\
\midrule
DeluluNet (ours) & 51.8 & 49.2 & 50.9 \\
$+$ Masking Token & \cellcolor{red!15}50.7 (-1.1) & \cellcolor{red!15}48.2 (-1.0) & \cellcolor{red!15}50.8 (-0.1) \\
$-$ Prefusion MSE & \cellcolor{red!15}49.2 (-2.5) & \cellcolor{red!15}48.3 (-0.9) & \cellcolor{red!15}50.6 (-0.3) \\
$-$ Latent MSE & \cellcolor{red!15}48.8 (-3.0) & \cellcolor{red!15}48.6 (-0.6) & \cellcolor{red!15}49.7 (-1.2) \\
$-$ Batch Mixing & \cellcolor{red!15}50.9 (-0.8) & \cellcolor{red!15}48.5 (-0.7) & \cellcolor{red!15}48.8 (-2.1) \\
\bottomrule
\end{tabular}}
        \captionof{table}{\textbf{DeluluNet Components Ablation on reBEN} ($\Ma$:S1, $\Mb$:S2), deltas relative to DeluluNet. 
        \emph{$+$Masking token config} uses per-modality learned mask tokens to replace masking transformers; 
        \emph{$-$Prefusion MSE config} removes direct supervision for masking transformers to hallucinate cross-modality representations; \emph{$-$Latent MSE config} removes latent reconstruction of the teacher; \emph{$-$Batch Mixing config} removes labeled batch learning. Each component improves its intended objective; all configurations are hyper-parameter tuned with 32 runs over the same search space.}
        \label{tab:ablation}
    \end{minipage}
    \hfill
    \begin{minipage}[t]{0.35\textwidth}
        \vspace{0pt}
        \centering
        \includegraphics[width=\textwidth]{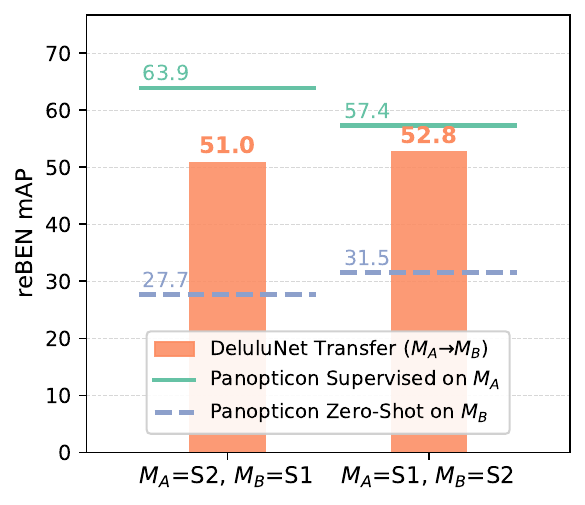}
        \captionof{figure}{\textbf{Panopticon Zero-Shot vs. Delulu for transfer.}
        DeluluNet takes advantage of widely available unlabeled paired multi-modal data, achieving significantly better transfer performance.}
    \label{fig:panopticon-vs-delulu}
    \end{minipage}
\end{figure}

\begin{figure}[t]
    \centering
        \begin{minipage}[c]{0.34\linewidth}
            \centering
            \includegraphics[width=\linewidth]{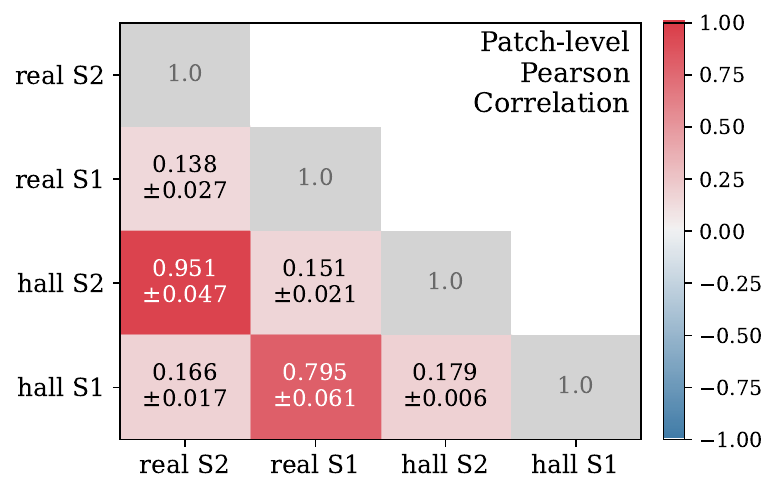}
        \end{minipage}
        \hfill
        \begin{minipage}[c]{0.63\linewidth}
            \centering
            \includegraphics[width=\linewidth]{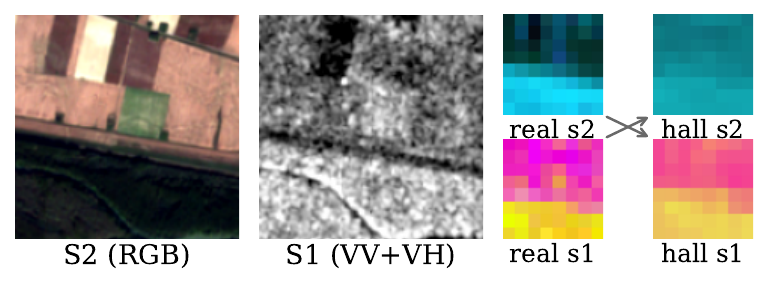}
        \end{minipage}
        \caption{
            \textbf{(left) Patch-level Pearson Correlation between real and hallucinated patch representations (quantitative).}
            Despite low correlation between real S1 and S2 tokens, masking transformers achieve high linear correlation between real and hallucinated patch tokens across both modalities.
            \textbf{(right) DeluluNet modality hallucinations (qualitative).}
            Top-3 PCA components visualized as RGB show high similarity between real and hallucinated feature maps for both S1$\rightarrow$S2 and S2$\rightarrow$S1.
        }
        \label{fig:masking-transformer-analysis}
    \vspace{2em}
    \centering
    \begin{minipage}{0.56\textwidth}
        \centering
        \resizebox{\textwidth}{!}{\begin{tabular}{cccccc}
\toprule
Model & Modality & Params (M) & GMACs & Max Batch Size & Throughput \\
\midrule
\multirow{2}{*}{DINOv3} & s1 (2) & 85.5 & 6.0 & 8192 & 1759 \\
 & s2 (12) & 87.4 & 6.1 & 8192 & 1719 \\
\midrule
\multirow{3}{*}{DeluluNet} & s1 (2) & 113.9 & 11.3 & 4096 & 959 \\
 & s2 (12) & 115.8 & 11.4 & 4096 & 952 \\
 & s1+s2 (14) & 109.1 & 12.2 & 4096 & 877 \\
\midrule
\multirow{3}{*}{Panopticon} & s1 (2) & 99.0 & 9.0 & 4096 & 1261 \\
 & s2 (12) & 99.0 & 18.0 & 512 & 660 \\
 & s1+s2 (14) & 99.0 & 19.8 & 512 & 607 \\
\bottomrule
\end{tabular}}
        \captionof{table}{\textbf{Computation cost comparison among uni-modal DINO, DeluluNet, and Panopticon.} DINOv3 is low in GMACs and high in throughput, but is uni-modal. DeluluNet is more expensive than Panopticon when hallucinating S2 (12 bands) from S1 (2 bands), but cheaper when hallucinating S1 from S2. Note that DeluluNet activates different numbers of parameters when in different mode due to modality-specific patch embedders and conditionally-activated masking transformers. Experiments performed on a single NVIDIA L40S GPU, GMACs is per-sample.}
        \label{tab:computation}
    \end{minipage}
    \hfill
    \begin{minipage}{0.4\textwidth}
        \centering
        \includegraphics[width=\textwidth]{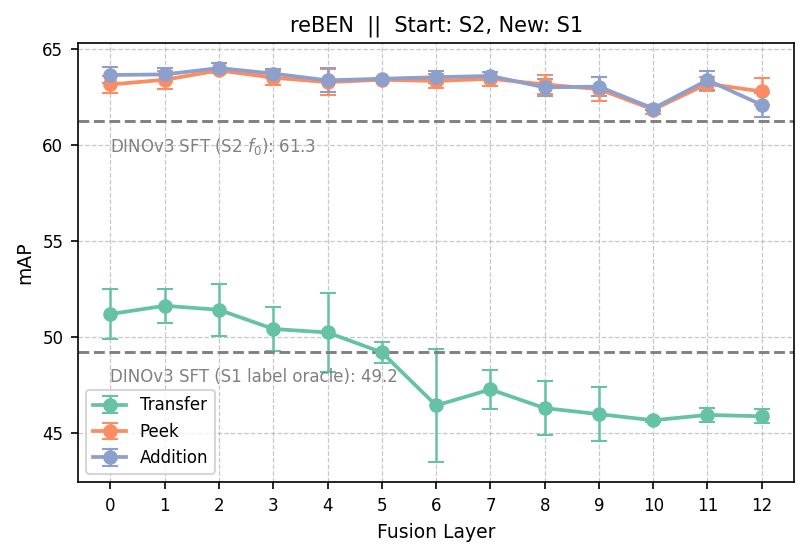}
        \captionof{figure}{\textbf{DeluluNet benefits from more allocation of layers for modality fusion}: earlier fusion tends to lead to better results across transfer, peeking, and addition. 
        For modality transfer, earlier fusion even leads to transfer results that surpass DINOv3 supervised fine-tuned on Sentinel-1 data.}
        \label{fig:fusion_time}
    \end{minipage}
\end{figure}

\subsection{Under The Hood of DeluluNet: Masking Transformer Analysis and Component Ablations}
\label{sec:masking_transformer_analysis}
We analyze three key design choices of DeluluNet—\emph{masking transformers}, \emph{batch mixing}, and \emph{auxiliary MSE loss terms}—to validate that each contributes to performance as intended.

\paragraph{Masking transformer analysis.}
Masking transformers facilitate cross-modality learning through modality hallucination during training, and enable DeluluNet to make accurate predictions by generating pseudo multi-modal representations even when only a subset of modalities is available during testing.
Figure \ref{fig:masking-transformer-analysis} shows that masking transformers are effective at imputing representations of missing modalities, measured quantitatively by Pearson correlation (left) and qualitatively by PCA visualization of real and hallucinated feature maps (right).
Masking transformers can hallucinate Sentinel-2 representations from Sentinel-1 input and vice versa, resulting in grounded pseudo multi-modal representations suitable for fusion.
Table~\ref{tab:ablation} shows that replacing masking transformers with per-modality mask tokens hurts DeluluNet on modality transfer and peeking tasks—the settings where masking transformers impute the missing modality.
This verifies that accurate hallucinations of multi-modal intermediate feature maps are beneficial for downstream performance.
Beyond hallucination quality, our ablation in Figure~\ref{fig:fusion_time} shows that an earlier fusion strategy is preferable to a late one, consistent with recent works in RSFMs \citep{tseng2025galileo,labatie2025maestromaskedautoencodersmultimodal}, and further verifies that DeluluNet performs meaningful fusion using hallucinated multi-modal representations.

\paragraph{Batch mixing: learning from real label with pseudo data and from pseudo label with real data.}
Standard distillation methods like KD and TTM learn from teacher predictions (pseudo labels) with real input data when no labels are available. 
DeluluNet additionally learns from the unimodal labeled data when the input modality is incomplete.
This is achievable for DeluluNet because of masking transformers' ability to impute missing modality with predicted intermediate feature maps, i.e. learn from real labels with pseudo inputs.
The two directions of batch mixing serve distinct purposes: unlabeled batches from $\Da$ ground the model in real multi-modal inputs, preventing it from overfitting to hallucinated representations; labeled batches from $\Dl$ provide clean ground truth supervision, preventing the model from overfitting to the teacher's prediction mistakes.
Table~\ref{tab:ablation} shows that batch mixing improves DeluluNet's performance across all three scenarios.

\paragraph{Loss ablations: Pre-fusion MSE and Latent MSE}
In addition to the two loss terms for prediction heads that enable batch mixing (distillation loss and supervised loss), DeluluNet receives pre-fusion MSE and latent MSE as supervision.
Prefusion MSE directly encourages masking transformers to impute the masked tokens/modality with predictions similar to the actual tokens. 
Latent MSE takes advantage of having a starting uni-modal $f_0$ that can extract useful latents from uni-modal $\Ma$ inputs, and is similar in construction to feature-based distillation methods \citep{heo2019comprehensive}.
Table~\ref{tab:ablation} shows that each term is crucial as intended: Pre-fusion MSE improves performance for transfer and peeking, when masking transformers are used to hallucinate the modality unavailable in inputs, and Latent MSE largely improves performance across the three scenarios.

\paragraph{Comparison to Panopticon zero-shot transfer.}
Unlike regular remote sensing models, a small group of sensor-agnostic models can extract features from unseen modalities; notably, Panopticon \citep{waldmann2025panopticon} can perform modality transfer zero-shot without paired unlabeled data.
However, such paired unlabeled data is abundant in remote sensing due to the continuous collection of orbiting satellites, and Figure~\ref{fig:panopticon-vs-delulu} shows that leveraging it allows DeluluNet to significantly outperform zero-shot transfer.

\paragraph{Computational cost of DeluluNet}
The amount of computation required by DeluluNet
depends on the changing modality scenario.
For modality peeking and transfer, where DeluluNet hallucinates multi-modal representation to perform pseudo-fusion, DeluluNet is generally more expensive than uni-modal foundation models like DINO due to this hallucination and token concatenation.
However, even computationally light multi-modal remote sensing models such as Panopticon \citep{waldmann2025panopticon} tend to incur higher 
computational cost than uni-modal RGB models.
This is due to design choices common in RSFMs such as per-band embedding \citep{waldmann2025panopticon}, channel-group concatenation \citep{cong2022satmae,tseng2025galileo}, and small patch size \citep{clay2025model,feng2025tesseratemporalembeddingssurface}.
In Table~\ref{tab:computation} we use Panopticon as an RSFM example to show that DeluluNet's computation is comparable for both uni-modal and multi-modal inputs.

\section{Discussion: Limitations and Extensions}
\label{sec:discussion}
DeluluNet demonstrates that a single unified framework can adapt existing uni-modal 
models to changing satellite modalities—without re-labeling—across transfer, 
addition, and peeking scenarios.
The current work establishes this in the two-modality setting, with experiments 
restricted to static land-use tasks and paired unlabeled adaptation data.
Below we discuss these limitations, and the broader opportunities DeluluNet opens up for the remote sensing community in future orbits.

\paragraph{Adapting foundation models to new sensor modalities.}
While we initialize DeluluNet from fine-tuned RGB DINOv3 to avoid remote sensing modality leakage in our experiments, in principle DeluluNet applies to adapting existing RSFMs to entirely new  and future sensor modalities.
As satellite constellations grow and sensors diversify, even the most capable RSFMs today will face modalities unseen during pretraining—and retraining from scratch each time is unsustainable.
DeluluNet offers a practical path forward: augmenting an existing fine-tuned RSFM with modular components and unlabeled paired data, rather than rebuilding it.
For practitioners wishing to work with sensor modalities not supported by existing RSFMs, DeluluNet offers an immediately deployable solution: rather than waiting for a new foundation model to be pretrained, one can augment an existing model with modular components using unlabeled paired data.

\paragraph{Temporal data and time-sensitive tasks.}
DeluluNet's paired adaptation split $\Da$ assumes that co-located observations from 
different satellites capture the same targets—a reasonable assumption for static 
land-use tasks, but one that breaks down for time-sensitive applications such as 
disaster monitoring \citep{revankar2025monitrs} or unlicensed fishing detection 
\citep{paolo2022xview3}.
In these settings, fast-shifting targets may no longer be present when a second 
satellite passes over the same region, introducing temporal misalignment that 
undermines reliable modality imputation.
Explicitly modeling the temporal offset between satellite captures—and its effect 
on cross-modal correspondence—is therefore an important direction for extending 
DeluluNet to dynamic remote sensing tasks.

\paragraph{Continual self-supervised learning.}
While DeluluNet adapts an existing task model to new modalities on demand, a 
complementary direction is to continually update the foundation model itself as 
new sensors come online—eliminating the need to retrain from scratch each time 
a new modality is added.
Remote sensing is uniquely suited for this: the continuous data collection nature 
of satellites means unlabeled paired observations accumulate naturally as new 
missions launch, providing a steady stream of data to update on without any 
additional annotation effort.

\paragraph{Toward \textit{ever}-changing modalities.}
The three scenarios of changing modalities in this work—transfer, addition, and peeking—provide the elements to describe how many modality changes can unfold over time.
While we restrict our experiments to the two-modality setting to first understand the minimal case, as new sensors are launched and old ones are retired, transfer, addition, and peeking can be applied in sequence, each building on the adapted model from the previous step.
DeluluNet shows a promising first step toward sustainable, relabeling-free deployment of remote sensing models subject to the reality of evolving satellite constellations.
We hope that our work motivates the community to develop long-term benchmarks covering diverse modalities and prediction tasks, and to design methods for adapting remote sensing models to change along with the sensors they rely on.

\begin{ack}
GP and ES are supported by Canada CIFAR AI Chairs.
We acknowledge the support of the Natural Sciences and Engineering Research Council of Canada (NSERC: RGPIN-2024-06405).
Resources used in preparing this research were provided, in part, by the Province of Ontario, the Government of Canada through CIFAR, and companies sponsoring the Vector Institute.
AF is primarily supported by an NSERC PGS-D scholarship.
We thank Arjun Rao and Seungyeon Baek for their helpful review and feedback on the experiments and exposition.
\end{ack}

\bibliographystyle{plainnat}
\bibliography{references}

\clearpage

\appendix

\section{Label-Free Early Stopping for DeluluNet Transfer and Addition}
\label{apx:validation}
Similar to training splits $\Dl$ and $\Da$, we assume a labeled uni-modal $\Ma,\mathcal{Y}$ validation set and an unlabeled multi-modal $\Ma,\Mb$ validation set.
Since peeking mode only takes in $\Ma$, checkpoint selection follows standard validation evaluation.
For \emph{transfer} and \emph{addition}, which operate on the unlabeled multimodal split, we instead use each mode's agreement with the supervised uni-modal $f_0$ as a proxy metric.

\begin{figure}[h]
    \centering
    \includegraphics[width=\textwidth]{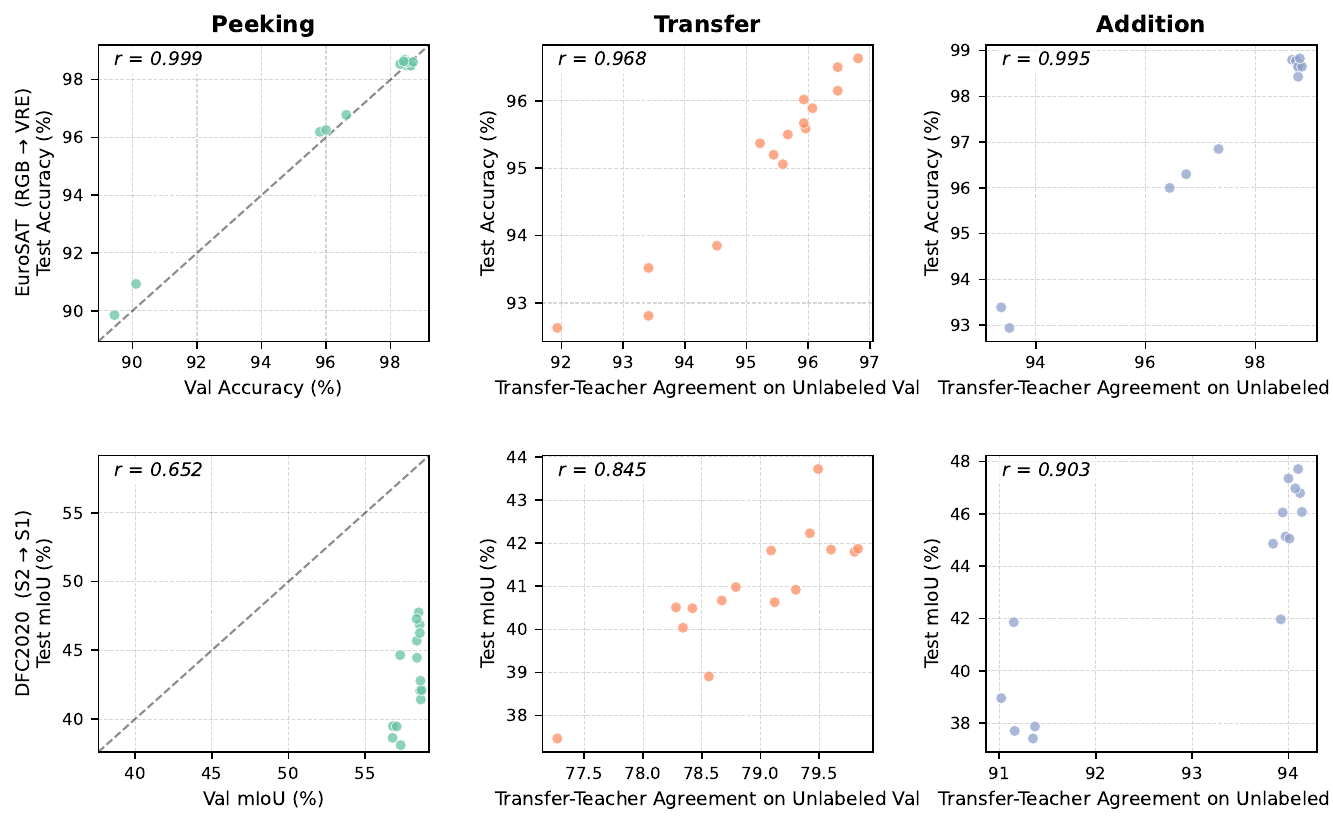}
    \caption{\textbf{Label-Free Validation Vs. Test metric} (left) peeking uses standard validation evaluation since we have labels for $\Ma$. (center\&right) Transfer and Addition uses their agreement with teacher $f_0$ times peeking performance as val metric, providing a validation score that doesn't require labels for multimodal validation split, and corresponds to test split metric reasonably well.}
    \label{fig:validation}
\end{figure}

There are many ways to define ``agreement", both hard (using peek prediction as pseudo labels) or ``soft" (with divergence or similarity metrics).
Figure~\ref{fig:validation} shows that this proxy correlates well with the true test metric when validation score is defined as top-1 agreement.
It is thus useful for both hyper-parameter tuning and early stopping without access to labels on the multimodal split.

\section{Full Cross-Band EuroSAT Results}
\label{apx:eurosat}
The main experiments in Section~\ref{sec:experiment} focus on cross-satellite modality changes, which are more practically relevant and challenging given the greater distinction between modalities.
Here we complement these with full cross-band results, evaluating transfer, addition, and peeking across the three major Sentinel-2 band groups from EuroSAT: RGB (B-02, B-03, B-04), VRE (B-05, B-06, B-07), and NIR (B-08, B-8A).
\begin{table}[t]
\centering
\caption{\textbf{Cross-Band Modality Transfer} results show that DeluluNet transfers more effectively than knowledge distillation (KD) and transformed teacher matching (TTM) for cross-band setting, at times even surpasses supervised fine-tuned DINOv3 (DINO SFT) on $M_B$.}
\label{tab:eurosat-transfer}
\resizebox{0.9\textwidth}{!}{
\begin{tabular}{cc | c c c c c c}
\toprule
  \multicolumn{2}{c|}{\shortstack[c]{Dataset\\(metric)}} & \multicolumn{6}{c}{\shortstack[c]{EuroSAT\\(Acc)}} \\
  \midrule
  \multicolumn{2}{c|}{Start ($M_A$)} & \multicolumn{2}{c}{RGB} & \multicolumn{2}{c}{VRE} & \multicolumn{2}{c}{NIR} \\
  \multicolumn{2}{c|}{Transfer ($M_B$)} & VRE & NIR & RGB & NIR & RGB & VRE \\
  \midrule
  $f_0(M_A)$ & $f_0$ & 98.6 & 98.6 & 90.7 & 90.7 & 80.9 & 80.9 \\
  \midrule
  \multirow{2}{*}{Baselines} & KD & 88.6$\pm$1.3 & 74.8$\pm$7.4 & 86.2$\pm$2.0 & 75.3$\pm$3.9 & 81.4$\pm$0.8 & 80.7$\pm$0.6 \\
   & TTM & 89.5$\pm$1.0 & 77.4$\pm$5.7 & 88.2$\pm$0.7 & 75.0$\pm$3.7 & 81.4$\pm$0.8 & 80.8$\pm$1.0 \\
  Ours & Delulu & \textbf{96.4$\pm$0.2} & \textbf{88.4$\pm$0.6} & \textbf{89.0$\pm$0.5} & \textbf{78.9$\pm$1.7} & \textbf{85.0$\pm$0.2} & \textbf{82.3$\pm$0.4} \\
  \midrule
  \shortstack[c]{Oracle ($M_B$)} & DINOv3 & \textcolor{gray}{90.7} & \textcolor{gray}{80.9} & \textcolor{gray}{98.6} & \textcolor{gray}{80.9} & \textcolor{gray}{98.6} & \textcolor{gray}{90.7} \\
\bottomrule
\end{tabular}
}

\bigskip

\centering
\caption{\textbf{Cross-Band Modality Peeking} results show that DeluluNet is able to learn from observations of unlabeled and paired $\Ma$ and $\Mb$ data to make better predictions on $\Ma$, consistently surpassing Semi-SL baseline and DINO-init $M_A$ $f_0$.}
\label{tab:eurosat-peeking}
\resizebox{0.9\textwidth}{!}{
\begin{tabular}{cc | c c c c c c}
\toprule
  \multicolumn{2}{c|}{\shortstack[c]{Dataset\\(metric)}} & \multicolumn{6}{c}{\shortstack[c]{EuroSAT\\(Acc)}} \\
  \midrule
  \multicolumn{2}{c|}{Start ($M_A$)} & \multicolumn{2}{c}{RGB} & \multicolumn{2}{c}{VRE} & \multicolumn{2}{c}{NIR} \\
  \multicolumn{2}{c|}{New ($M_B$)} & VRE & NIR & RGB & NIR & RGB & VRE \\
  \midrule
  $f_0(M_A)$ & $f_0$ & 98.6 & 98.6 & 90.7 & 90.7 & 80.9 & 80.9 \\
  Baselines & MixMatch & \textbf{99.0$\pm$0.1} & \textbf{99.0$\pm$0.1} & 87.9$\pm$4.1 & 87.9$\pm$4.1 & 66.7$\pm$5.7 & 66.7$\pm$5.7 \\
  Ours & Delulu & 98.5$\pm$0.1 & 98.5$\pm$0.0 & \textbf{91.6$\pm$0.3} & \textbf{91.6$\pm$0.2} & \textbf{82.8$\pm$0.3} & \textbf{82.3$\pm$0.4} \\
\bottomrule
\end{tabular}
}

\bigskip

\centering
\caption{\textbf{Cross-Band Modality Addition} results show that DeluluNet is able to extract and fuse inputs from both $\Ma$ and $\Mb$ through modality addition.}
\label{tab:eurosat-addition}
\resizebox{0.9\textwidth}{!}{
\begin{tabular}{cc | c c c c c c}
\toprule
  \multicolumn{2}{c|}{\shortstack[c]{Dataset\\(metric)}} & \multicolumn{6}{c}{\shortstack[c]{EuroSAT\\(Acc)}} \\
  \midrule
  \multicolumn{2}{c|}{Start ($M_A$)} & \multicolumn{2}{c}{RGB} & \multicolumn{2}{c}{VRE} & \multicolumn{2}{c}{NIR} \\
  \multicolumn{2}{c|}{New ($M_B$)} & VRE & NIR & RGB & NIR & RGB & VRE \\
  \midrule
  $f_0(M_A)$ & $f_0$ & 98.6 & 98.6 & 90.7 & 90.7 & 80.9 & 80.9 \\
  Baselines & MKE & 97.5$\pm$0.9 & 97.9$\pm$0.5 & 90.2$\pm$0.2 & 89.4$\pm$0.3 & 77.4$\pm$4.0 & 77.0$\pm$2.0 \\
  Ours & Delulu & \textbf{98.7$\pm$0.1} & \textbf{98.7$\pm$0.1} & \textbf{91.5$\pm$0.1} & \textbf{91.6$\pm$0.1} & \textbf{82.3$\pm$0.3} & \textbf{81.9$\pm$0.2} \\
  \midrule
  \shortstack[c]{Oracle\\($M_A$+$M_B$)} & DINOv3 & \textcolor{gray}{98.6} & \textcolor{gray}{98.8} & \textcolor{gray}{98.6} & \textcolor{gray}{88.7} & \textcolor{gray}{98.8} & \textcolor{gray}{88.7} \\
\bottomrule
\end{tabular}
}
\end{table}

Tables~\ref{tab:eurosat-transfer}, \ref{tab:eurosat-peeking}, \ref{tab:eurosat-addition} confirm that findings from the $\Ma=$RGB, $\Mb=$VRE setting in Section~\ref{sec:experiment} generalize across Sentinel-2 sub-bands, and that DeluluNet serves as a reliable framework for both cross-satellite and cross-band transfer, peeking, and addition.

\section{DeluluNet Architecture Details}
\label{apx:architecture}
\paragraph{Modality Specific Feature Extractor.}
In Vision Transformers (ViT) \citep{dosovitskiy2021an}, image inputs are divided into patches and projected into token embeddings via a 2D convolution layer, where \texttt{in\_chan} is set to 3 for RGB inputs.
Since remote sensing modalities have varying numbers of input channels, it is common to use modality-specific patch embedders \citep{tseng2025galileo,herzog2025olmoearthstablelatentimage}.
DeluluNet follows this design: each modality receives a dedicated 2D convolution layer for patch embedding, followed by 3 transformer blocks where tokens from the same modality attend to each other.

\paragraph{Masking Transformers.}
Masking transformers are lightweight 2-layer transformers consisting of a standard self-attention layer followed by a custom cross-attention layer.
The cross-attention layer contains per-modality queries for $\Ma$ and $\Mb$, each of dimension $768$ and length 2—one for the CLS token and one for the patch token.
At forward time, the patch query is broadcast into $n$ copies, where $n = (\texttt{input-size} / \texttt{patch-size})^2$ is the number of tokens for that modality.
Each broadcast patch query is then augmented with ROPE positional encoding shared with the rest of DeluluNet, ensuring a principled and parameter-efficient design.

\paragraph{Modality Embeddings and Fusion Blocks.}
Modality embeddings encode modality information, since naive transformer layers treat all tokens equally and are blind to which modality a given token belongs to.
These are learned $d=768$ vectors added directly to intermediate representations from modality-specific feature extractors (or hallucinated ones from masking transformers).
The Fusion Transformer consists of 9 self-attention blocks; together with the 3-layer modality-specific transformers, this matches the ViT-Base architecture, which we initialize with DINOv3-Base weights.

\paragraph{Predictor Heads.}
We use simple linear heads across all evaluated tasks.
For DeluluNets, separate heads predict from the CLS token (for classification) or patch tokens (for segmentation) from both $\Ma$ and $\Mb$.
This resembles Shallow Ensemble \citep{lee2015m}, where multiple heads predict from the same backbone—with the key difference that DeluluNet's heads operate on features from different modalities.
Although this potentially introduces an implicit asymmetry in predictive power when some modality features are hallucinated by masking transformers, joint optimization of the predictor heads during training \citep{zhou2025asymmetric} mitigates this concern.

\paragraph{Train-Time-Only Component.}
DeluluNet includes a lightweight one-layer transformer that predicts the $f_0$ representation from fused $\Ma$ representations.
This provides flexibility in what DeluluNet learns, rather than forcing it to directly match the teacher's representation.



\end{document}